\documentclass[a4paper,twoside]{article}

\usepackage{epsfig}
\usepackage{subcaption}
\usepackage{calc}
\usepackage{amssymb}
\usepackage{amstext}
\usepackage{amsmath}
\usepackage{amsthm}
\usepackage{multicol}
\usepackage{pslatex}
\usepackage{url} 
\usepackage{url}

\usepackage{apalike}
\usepackage{algorithm2e}
\usepackage[bottom]{footmisc}
\usepackage{SCITEPRESS}     

\begin{document}

\title{Rethinking Retrieval: From Traditional Retrieval Augmented Generation to Agentic and Non-Vector Reasoning Systems in the Financial Domain for Large Language Models}

\author{
 Elias Lumer, 
 Matt Melich,
 Olivia Zino, 
 Elena Kim,
 Sara Dieter,\\
 Pradeep Honaganahalli Basavaraju,
 Vamse Kumar Subbiah,\\
 James A. Burke and
 Roberto Hernandez
 \\
 \textit{\\PricewaterhouseCoopers U.S.}
}


\abstract{
Recent advancements in Retrieval-Augmented Generation (RAG) have enabled Large Language Models to answer financial questions using external knowledge bases of U.S. SEC filings, earnings reports, and regulatory documents. However, existing work lacks systematic comparison of vector-based and non-vector RAG architectures for financial documents, and the empirical impact of advanced RAG techniques on retrieval accuracy, answer quality, latency, and cost remain unclear. We present the first systematic evaluation comparing vector-based agentic RAG using hybrid search and metadata filtering against hierarchical node-based systems that traverse document structure without embeddings. We evaluate two enhancement techniques applied to the vector-based architecture, i) cross-encoder reranking for retrieval precision, and ii) small-to-big chunk retrieval for context completeness. Across 1,200 SEC 10-K, 10-Q, and 8-K filings on a 150-question benchmark, we measure retrieval metrics (MRR, Recall@5), answer quality through LLM-as-a-judge pairwise comparisons, latency, and preprocessing costs. Vector-based agentic RAG achieves a 68\% win rate over hierarchical node-based systems with comparable latency (5.2 compared to 5.98 seconds). Cross-encoder reranking achieves a 59\% absolute improvement at optimal parameters (10, 5) for MRR@5. Small-to-big retrieval achieves a 65\% win rate over baseline chunking with only 0.2 seconds additional latency. Our findings reveal that applying advanced RAG techniques to financial Q\&A systems improves retrieval accuracy, answer quality, and has cost-performance tradeoffs to be considered in production.}

\maketitle
\normalsize
\setcounter{footnote}{0}

\section{\uppercase{Introduction}}\label{sec:Introduction}

Recent advancements in Large Language Models (LLMs) have enabled powerful question answering systems that leverage external knowledge bases through Retrieval-Augmented Generation (RAG) \cite{lewis2020rag,karpukhin2020dpr}. These systems retrieve relevant information from vector databases and inject context into the model's prompt at inference time, improving factual accuracy and reducing hallucinations \cite{gao2023rag_survey,ni2024rag_survey}. However, naïve RAG is limited in addressing complex multi-hop queries, motivating advanced RAG strategies including chunking optimization \cite{chiang2024_optimizing_rag}, query transformation \cite{ma2023query_rewriting,anthropic2024contextual_retrieval}, and reranking \cite{nogueira2019_passage_rerank,sun2023_llm_reranking,yan2024_corrective_rag}.

A significant gap remains in understanding how vector-based RAG systems compare to emerging non-vector reasoning approaches for financial documents \cite{gao2023rag_survey}. Hierarchical node-based reasoning systems organize documents through structured table-of-contents representations, enabling traversal-based retrieval without dense embeddings \cite{vectifyai_pageindex_repo,vectifyai_pageindex_financebench}. Their effectiveness relative to semantic vector search remains unclear for complex financial queries requiring multi-hop reasoning across lengthy regulatory filings \cite{islam2023_financebench,chen2021_finqa,zhu2021_tatqa}.

Limited research exists on advanced RAG techniques for financial document question answering. While techniques such as cross-encoder reranking and small-to-big chunk retrieval improve general-domain QA \cite{nogueira2019_passage_rerank,sun2023_llm_reranking,chiang2024_optimizing_rag}, financial QA systems lack systematic evaluation across retrieval accuracy, answer quality, latency, and cost trade-offs \cite{setty2024_financial_rag_retrieval,michel2025_fincare,wu2025_finsage}. A gap remains in controlled comparisons isolating the impact of individual enhancement techniques on end-to-end system performance.

In this paper, we present the first systematic evaluation comparing vector-based agentic RAG against hierarchical node-based reasoning systems for financial document question answering. Our vector-based baseline consists of an agentic RAG system using hybrid search with metadata filtering \cite{anthropic2024contextual_retrieval}, corrective RAG \cite{yan2024_corrective_rag}, and standard token-based chunking. We compare this against a hierarchical node-based reasoning system that traverses documents through structured table-of-contents representations without vector embeddings. Additionally, we evaluate two advanced enhancement techniques applied independently to the vector-based architecture, including cross-encoder reranking to improve retrieval precision through fine-grained relevance scoring, and small-to-big retrieval to augment target chunks with neighboring context for comprehensive information coverage. For cross-encoder reranking, we systematically evaluate various parameter configurations (initial retrieval depth $k_{initial}$ and final selection size $k_{final}$), comparing reranked retrieval against baseline vector retrieval. For small-to-big retrieval, we compare expanded context provision against standard chunk retrieval. We conduct experiments across 150 questions spanning 10-K and 10-Q filings, measuring retrieval metrics (Mean Reciprocal Rank, Recall@5), answer quality through LLM-as-a-judge evaluation \cite{zheng2023_llm_as_judge}, latency, and preprocessing costs.

After evaluating both retrieval architectures and advanced RAG strategies across 1,200 SEC filings (10-K, 10-Q, and 8-K) on our 150-question financial document benchmark, we find that vector-based agentic RAG achieves a 68\% win rate over hierarchical node-based systems with comparable latency (5.2 compared to 5.98 seconds). We evaluate two enhancement techniques applied to the vector-based architecture: cross-encoder reranking achieves a 59\% absolute improvement at optimal parameters (10, 5) for MRR@5, improving Mean Reciprocal Rank from 0.160 to 0.750 with perfect Recall@5 (1.00). Small-to-big retrieval achieves a 65\% win rate over baseline chunking with only 0.2 seconds additional latency. Our findings reveal that applying advanced RAG techniques to financial question answering systems improves retrieval accuracy and answer quality, with cost-performance tradeoffs to be considered in production deployment.

\section{\uppercase{Related Work}}\label{sec:relatedwork}

\subsection{Retrieval-Augmented Generation}
Retrieval-Augmented Generation (RAG) enhances Large Language Models with external knowledge, improving factual accuracy and reducing hallucinations \cite{lewis2020rag}. Dense passage retrieval methods leverage bi-encoders to map queries and documents into shared embedding spaces \cite{karpukhin2020dpr}. Traditional RAG systems retrieve the top-k most relevant documents based on embedding distance \cite{manning2008ir}. However, naïve RAG faces limitations in handling multi-hop queries and long documents, motivating advanced enhancement techniques \cite{gao2023rag_survey,singh2025agentic_rag_survey}.

\subsection{Advanced RAG Enhancement Techniques}
Advanced RAG methods enhance naive RAG through pre-retrieval, intra-retrieval, and post-retrieval strategies \cite{gao2023rag_survey,ni2024rag_survey}. Pre-retrieval methods include small-to-big chunking where smaller chunks are retrieved but larger surrounding context is provided to the model \cite{chiang2024_optimizing_rag,anthropic2024contextual_retrieval}. The "lost in the middle" phenomenon demonstrates that models struggle to access information in long contexts \cite{liu2023_lost_in_middle}. Intra-retrieval techniques include query rewriting \cite{ma2023query_rewriting} and metadata filtering \cite{anthropic2024contextual_retrieval,dadopoulos2025metadatadrivenretrievalaugmentedgenerationfinancial,lumer2025tooltoagentretrievalbridgingtools}. Post-retrieval processes include reranking with cross-encoder models \cite{nogueira2019_passage_rerank,sun2023_llm_reranking} and corrective retrieval strategies \cite{yan2024_corrective_rag,asai2023selfrag}. Agentic RAG systems equip LLM agents with retrieval tools, enabling dynamic query formulation \cite{singh2025agentic_rag_survey,lumer2025memtooloptimizingshorttermmemory,lumer2024toolshedscaletoolequippedagents}. Hybrid retrieval combines semantic vector search with lexical methods like BM25 \cite{robertson2009_bm25}.

\subsection{Non-Vector and Structured Retrieval Approaches}
Index-free RAG approaches leverage structured representations such as hierarchical document structures to enable reasoning-based retrieval without vector embeddings \cite{openai2025_index_free_rag,openai2024_rag_help}. Hierarchical node-based reasoning systems organize documents into tree structures where nodes represent topical sections and LLMs traverse the hierarchy to locate relevant content \cite{vectifyai_pageindex_financebench}. While these non-vector approaches reduce preprocessing costs, their effectiveness relative to semantic vector search remains unclear for complex queries \cite{vectifyai_pageindex_repo,openai2025_index_free_rag}.

\subsection{Financial Document Question Answering}
Financial document question answering presents unique challenges due to lengthy regulatory filings and domain-specific terminology \cite{islam2023_financebench}. SEC Form 10-K annual reports span 100-300 pages, Form 10-Q quarterly reports provide 30-80 page interim updates, and Form 8-K reports disclose material events \cite{sec_form10k,sec_form10q,sec_form8k}. Benchmarks include FinQA \cite{chen2021_finqa}, TAT-QA \cite{zhu2021_tatqa}, and FinanceBench \cite{islam2023_financebench}. Recent work explores RAG-based systems emphasizing preprocessing strategies \cite{setty2024_financial_rag_retrieval} and metadata-driven retrieval \cite{dadopoulos2025metadatadrivenretrievalaugmentedgenerationfinancial,lumer2025scalemcpdynamicautosynchronizingmodel}. However, existing work does not systematically compare vector-based, non-vector RAG architectures, and modern advanced RAG strategies across retrieval accuracy, answer quality, latency, and cost trade-offs.

\begin{figure}[t]
\small
\begin{verbatim}
{
  "doc_name": "2023-annual-report.pdf",
  "structure": [
    {
      "title": "Monetary Policy and Developments",
      "start_index": 9,
      "end_index": 9,
      "nodes": [
        {
          "title": "March 2024 Summary",
          "start_index": 9,
          "end_index": 14,
          "node_id": "0004"
        },
        {
          "title": "June 2023 Summary",
          "start_index": 15,
          "end_index": 20,
          "node_id": "0005"
        }
      ],
      "node_id": "0003"
    },
    {
      "title": "Financial Stability",
      "start_index": 21,
      "end_index": 21,
      "nodes": [
        {
          "title": "Monitoring Vulnerabilities",
          "start_index": 22,
          "end_index": 28,
          "node_id": "0007"
        }
      ],
      "node_id": "0006"
    }
  ]
}
\end{verbatim}
\caption{Example hierarchical node tree structure for a Federal Reserve annual report. Each node contains a title, page range (start\_index, end\_index), and unique identifier, with nested nodes representing document subsections.}
\label{fig:node_tree_structure}
\end{figure}

\section{\uppercase{Method}}

\subsection{Dataset Construction and Document Processing}\label{subsec:dataset_construction}

\subsubsection{Document Corpus}

Our evaluation corpus consists of 1,200 U.S. SEC filings (10-K, 10-Q, and 8-K) from Fortune 500 companies (2020-2025) \cite{sec_form10k,sec_form10q,sec_form8k}. Documents average 73,175 tokens, reflecting real-world complexity \cite{islam2023_financebench,setty2024_financial_rag_retrieval}.

\subsubsection{Question-Answer Pair Generation}

We created 150 question-answer pairs balanced across three complexity categories: 65 multi-hop questions, 65 single-hop questions, and 20 summary questions \cite{islam2023_financebench,chen2021_finqa}. Each question was manually annotated with ground-truth answers and page locations. We constructed specialized subsets of 75 questions for architecture comparison, 50 questions for cross-encoder reranking evaluation, and 50 questions for small-to-big retrieval assessment.

\subsubsection{Hierarchical Node Tree Generation}

For each document, we generated hierarchical node tree representations \cite{vectifyai_pageindex_repo,vectifyai_pageindex_financebench} where each node represents a topical section with associated page ranges. We evaluated three models: OpenAI GPT-4o, Google Gemini 2.5 Flash, and OpenAI GPT-4.1 mini. Figure \ref{fig:node_tree_structure} shows an example node tree structure. We assessed preprocessing costs with and without node-level summaries, selecting OpenAI GPT-4o for final generation based on structure quality and computational efficiency.

\subsection{Retrieval System Architectures}\label{subsec:retrieval_architectures}

We compare two foundational retrieval architectures for financial document question answering, representing distinct paradigms for accessing relevant context from lengthy regulatory filings.

\subsubsection{Vector-Based Agentic RAG Using Hybrid Search and Metadata Filtering}

Our baseline vector-based system implements an agentic RAG architecture combining hybrid retrieval with metadata filtering \cite{karpukhin2020dpr,anthropic2024contextual_retrieval,singh2025agentic_rag_survey}. Documents are chunked using 512-token chunks with 50-token overlap. Each chunk is embedded using OpenAI text-embedding-ada-002 and stored in Azure AI Search with metadata. At query time, the LLM agent formulates search queries and retrieves the top-k most relevant chunks through hybrid search combining semantic and lexical matching \cite{lewis2020rag,gao2023rag_survey}.

\subsubsection{Hierarchical Node-Based Reasoning System}

The hierarchical node-based reasoning system organizes documents through structured table-of-contents representations, enabling traversal-based retrieval without vector embeddings \cite{vectifyai_pageindex_repo,vectifyai_pageindex_financebench}. Each document is represented as a hierarchical node tree where nodes correspond to document sections with page ranges (Figure \ref{fig:node_tree_structure}). At query time, the LLM traverses the hierarchy by selecting relevant nodes, then retrieves corresponding page ranges as context. This approach eliminates embedding generation costs but relies on the LLM's reasoning capabilities to navigate document structure.

\subsection{Advanced RAG Enhancement Techniques}\label{subsec:enhancement_techniques}

We evaluate two advanced RAG techniques independently: cross-encoder reranking for improving retrieval precision, and small-to-big retrieval for enhancing context quality. Each technique is assessed by comparing performance against baseline vector retrieval.

\subsubsection{Cross-Encoder Reranking}

Cross-encoder reranking refines initial vector retrieval by reordering chunks based on fine-grained relevance scoring \cite{nogueira2019_passage_rerank,sun2023_llm_reranking}. The system retrieves $k_{initial}$ chunks through vector search, then a cross-encoder model scores each chunk by jointly encoding the query-chunk pair. The system selects the top $k_{final}$ chunks based on cross-encoder scores, where $k_{final} \leq k_{initial}$. We test parameter settings ranging from ($k_{initial}=10$, $k_{final}=5$) to ($k_{initial}=100$, $k_{final}=30$), measuring impact on retrieval accuracy, latency, and answer quality.

\subsubsection{Small-to-Big Retrieval}

Small-to-big retrieval addresses the tension between retrieval precision and context completeness \cite{chiang2024_optimizing_rag,anthropic2024contextual_retrieval}. Our implementation retrieves target chunks through vector search, then augments each chunk with its immediate neighbors (preceding and following chunks). This expansion preserves retrieval accuracy while supplying richer context. We compare synchronous and asynchronous implementation strategies, measuring answer quality improvements, latency overhead, and cost impact.

\begin{figure}[t]
\centering
\includegraphics[width=1\columnwidth]{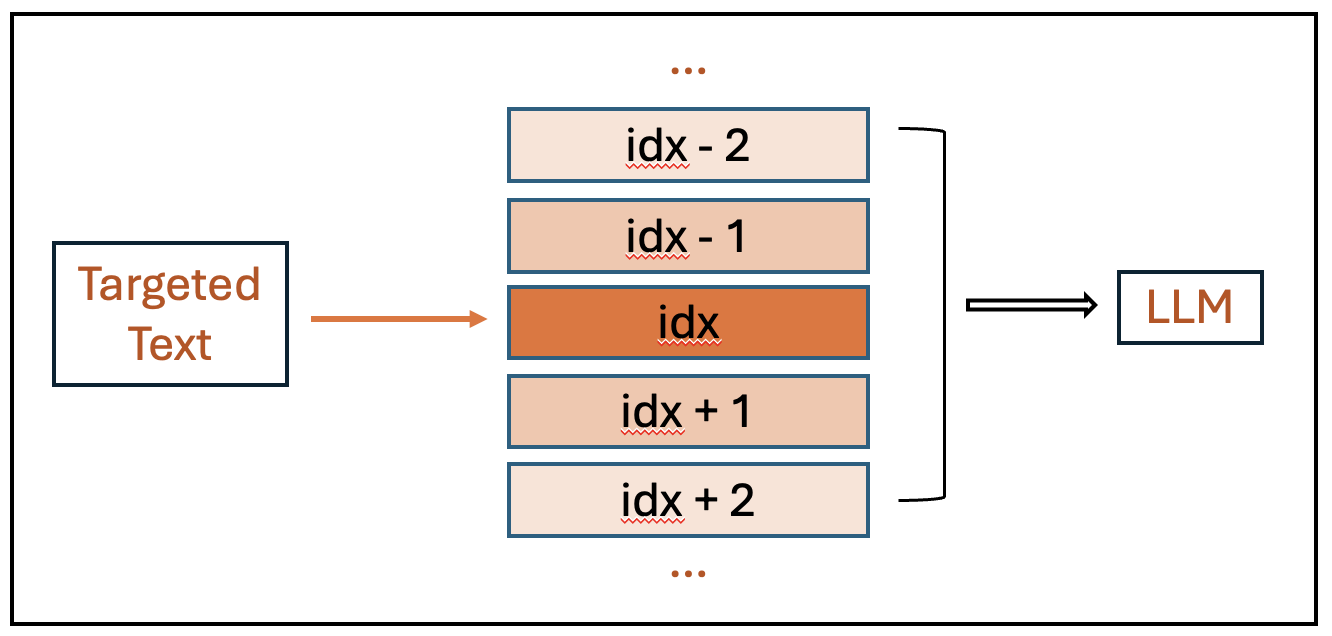}
\caption{Small-to-big retrieval strategy. The system retrieves a target chunk at index $idx$ through vector search, then expands context by including neighboring chunks ($idx-2$, $idx-1$, $idx+1$, $idx+2$) before providing the combined context to the LLM for answer generation.}
\label{fig:smalltobig}
\end{figure}

\subsection{Evaluation Framework}\label{subsec:evaluation_framework}

\subsubsection{Retrieval Metrics}

Retrieval performance is measured using Mean Reciprocal Rank (MRR) and Recall@5 \cite{manning2008ir,robertson2009_bm25}. MRR computes the average reciprocal rank of the first relevant chunk: $\text{MRR} = \frac{1}{|Q|} \sum_{i=1}^{|Q|} \frac{1}{\text{rank}_i}$. Recall@5 measures the fraction of relevant chunks retrieved within the top-5 results. Ground truth relevance is determined by annotated page numbers.

\subsubsection{Answer Quality Assessment}

Answer quality is evaluated using LLM-as-a-judge with pairwise comparisons \cite{zheng2023_llm_as_judge,huang2024_llm_as_judge_survey}. Anthropic Claude 4.5 Sonnet evaluates answer pairs across six criteria: accuracy, completeness, clarity, conciseness, relevance, and style. Win rates are computed as the proportion of pairwise comparisons where each system was preferred.

\subsubsection{Efficiency and Cost Analysis}

Performance efficiency is assessed through latency measurements and cost analysis \cite{setty2024_financial_rag_retrieval,wang2023_benchmarking_rag}. Latency is measured end-to-end from query submission to answer generation. Preprocessing costs encompass node tree generation for hierarchical systems and embedding generation for vector-based systems. Runtime costs include retrieval, reranking, and LLM inference.

\section{\uppercase{Experiments}}\label{sec:experiments}

\subsection{Experimental Settings}

\subsubsection{Dataset and Models}

Our evaluation uses the 150-question benchmark described in \ref{subsec:dataset_construction}, spanning 1,200 SEC filings (2020-2025). Questions include 65 multi-hop queries, 65 single-hop queries, and 20 summary queries. We construct specialized subsets of 75 questions for architecture comparison, 50 questions for cross-encoder reranking evaluation, and 50 questions for small-to-big retrieval assessment.

All experiments use OpenAI GPT-4o for answer generation, OpenAI text-embedding-ada-002 for embeddings, and Cohere \texttt{rerank-english-v3.0} for reranking \cite{cohere2024_rerank}. The vector database is Azure AI Search \cite{microsoft2024_ai_search}.

\begin{table}[t]
\centering
\caption{Preprocessing cost comparison for one company's SEC Form 10-Q and Form-10K different language models for hierarchical node tree generation. Token counts and model costs are reported for average 10-Q and 10-K filings.}
\label{tab:preprocessing_comparison}
\footnotesize
\begin{tabular}{l r r}
\hline
\textbf{Metric} & \textbf{Avg 10-Q} & \textbf{Avg 10-K} \\
\hline
Total Tokens & 2,149,038 & 3,336,665 \\
OpenAI GPT-4o Cost & \$7.21 & \$30.62 \\
Google Gemini 2.5 Flash Cost & \$0.99 & \$5.94 \\
OpenAI GPT-4.1 mini Cost & \$4.23 & \$22.27 \\
\hline
\end{tabular}
\end{table}

\begin{table}[t]
\centering
\caption{Preprocessing performance with node-level summary inclusion. Metrics show increased token consumption and cost but richer contextual information for hierarchical traversal. Model costs shown for OpenAI GPT-4o, Google Gemini 2.5 Flash, and OpenAI GPT-4.1 mini.}
\label{tab:with_summary}
\small
\begin{tabular}{l r r}
\hline
\textbf{Metric} & \textbf{Avg 10-Q} & \textbf{Avg 10-K} \\
\hline
Total Tokens & 126,414 & 667,333 \\
Input tokens & 117,115 & 627,381 \\
Output tokens & 9,299 & 399,952 \\
Latency (s) & 48.46 & 144.61 \\
GPT-4o Cost & \$0.42 & \$6.12 \\
Gemini 2.5 Flash Cost & \$0.06 & \$1.19 \\
GPT-4.1 mini Cost & \$0.25 & \$4.45 \\
\hline
\end{tabular}
\end{table}

\begin{table}[t]
\centering
\caption{Preprocessing performance without summary inclusion. Lower token consumption and cost but less contextual richness for node traversal reasoning. Model costs shown for OpenAI GPT-4o, Google Gemini 2.5 Flash, and OpenAI GPT-4.1 mini.}
\label{tab:without_summary}
\small
\begin{tabular}{l r r}
\hline
\textbf{Metric} & \textbf{Avg 10-Q} & \textbf{Avg 10-K} \\
\hline
Total Tokens & 56,833 & 299,622 \\
Input tokens & 52,080 & 281,536 \\
Output tokens & 4,753 & 18,086 \\
Latency (s) & 44.18 & 125.95 \\
GPT-4o Cost & \$0.20 & \$0.97 \\
Gemini 2.5 Flash Cost & \$0.03 & \$0.13 \\
GPT-4.1 mini Cost & \$0.11 & \$0.71 \\
\hline
\end{tabular}
\end{table}

\subsubsection{Evaluation Metrics}

We measure MRR, Recall@5, answer quality via LLM-as-a-judge \cite{zheng2023_llm_as_judge}, latency, and cost.

\subsection{Preprocessing and Node Generation}

\subsubsection{Model Comparison for Hierarchical Node Tree Generation}

Table \ref{tab:preprocessing_comparison} compares preprocessing costs across three models for processing a single company's filings. OpenAI GPT-4o produced superior structure coherence, while Gemini 2.5 Flash exhibited compatibility issues.

\begin{table*}[t]
\centering
\caption{Cross-encoder reranking performance across parameter settings compared to baseline vector retrieval without reranking. Bold indicates optimal parameter setting balancing MRR and latency. The optimal (10, 5) setting achieves MRR@5 of 0.750 compared to baseline of 0.160, representing an absolute improvement of 59\%.}
\label{tab:reranking_results}
\footnotesize
\begin{tabular}{c c c c}
\hline
\textbf{($k_{initial}$, $k_{final}$)} & \textbf{MRR@5} & \textbf{Recall@5} & \textbf{Avg Latency (s)} \\
\hline
Baseline (No Reranking) & 0.160 & 0.50 & 0.22 \\
\hline
(100, 20) & 0.519 & 1.00 & 6.01 \\
(100, 30) & 0.519 & 1.00 & 5.33 \\
(75, 25) & 0.536 & 1.00 & 4.52 \\
(75, 15) & 0.536 & 1.00 & 3.03 \\
(50, 10) & 0.550 & 1.00 & 4.15 \\
\textbf{(10, 5)} & \textbf{0.750} & \textbf{1.00} & \textbf{2.02} \\
(50, 5) & 0.550 & 1.00 & 2.52 \\
(20, 10) & 0.566 & 1.00 & 1.61 \\
(20, 5) & 0.479 & 1.00 & 2.13 \\
(10, 10) & 0.625 & 1.00 & 1.24 \\
\hline
\end{tabular}
\end{table*}

\subsubsection{Impact of Summary Inclusion}

Tables \ref{tab:with_summary} and \ref{tab:without_summary} show that summary inclusion increased 10-K preprocessing costs from \$0.97 to \$6.12 but provided richer node descriptions.

\subsection{Retrieval Architecture Comparison}

The vector-based agentic RAG system achieved a 68\% win rate over the hierarchical node-based system with faster latency (5.2 vs 5.98 seconds). The node-based system faced context window limitations, failing to answer 2 questions and providing 2 incorrect responses \cite{liu2023_lost_in_middle}. The vector-based system successfully retrieved relevant context for all questions.

\subsection{Cross-Encoder Reranking Evaluation}

\subsubsection{Results}

Table \ref{tab:reranking_results} shows reranking performance across parameter settings. Baseline vector retrieval achieved MRR@5 of 0.160 with Recall@5 of 0.50 and 0.22 second latency. The optimal ($k_{initial}=10$, $k_{final}=5$) setting achieved MRR@5 of 0.750 (59\% absolute improvement) with perfect Recall@5 of 1.00 and 2.02 second latency. All reranking settings achieved perfect recall (1.00) compared to baseline recall of 0.50.

\subsection{Small-to-Big Retrieval Evaluation}

Small-to-big retrieval achieved a 65\% win rate with 0.2 seconds additional latency. Per-query cost remained \$0.000078. Asynchronous retrieval (0.17 seconds) outperformed synchronous calls (0.34 seconds) by fetching neighboring chunks in parallel.

\section{\uppercase{Discussion}}\label{sec:discussion}

\subsection{Vector-Based and Hierarchical Reasoning Systems}

The vector-based agentic RAG system's 68\% win rate demonstrates that semantic embedding retrieval outperforms structured traversal for complex financial queries. The node-based system's context window constraints led to 2 failed answers and 2 incorrect responses, while the vector-based system successfully retrieved relevant context for all questions.

A key limitation of the hierarchical node-based approach was the retrieval bottleneck at the table-of-contents level. The LLM's ability to select relevant sections from the document structure proved less effective than vector-based semantic matching, suggesting either that the LLM struggled with hierarchical navigation for complex financial queries, or that our question set emphasized specific factual retrieval rather than broad document summarization where hierarchical traversal might excel. A promising direction for future work would be to combine vector search with hierarchical reasoning by using vector embeddings to narrow the initial scope of candidate nodes before LLM traversal, potentially addressing the table-of-contents selection bottleneck while preserving the structured reasoning benefits of hierarchical approaches.

Despite comparable latency (5.2 vs 5.98 seconds), the systems exhibit different cost structures. Node-based systems incur \$30.62 per 10-K filing for preprocessing. However, the vector-based system's superior answer quality justifies the infrastructure investment for accuracy-critical applications \cite{karpukhin2020dpr,lewis2020rag,singh2025agentic_rag_survey}.

For hierarchical node tree generation, OpenAI GPT-4o produced superior structure coherence compared to GPT-4.1 mini, while Gemini 2.5 Flash exhibited compatibility issues. Summary inclusion increased preprocessing costs 6.3× (from \$0.97 to \$6.12 per 10-K for GPT-4o) but provided richer contextual information (Tables \ref{tab:with_summary} and \ref{tab:without_summary}). Selective summary inclusion based on query complexity may optimize cost-performance balance \cite{izacard2021fid}.

\subsection{Cross-Encoder Reranking}

Cross-encoder reranking substantially improved retrieval quality. Baseline vector retrieval achieved MRR@5 of 0.160 with Recall@5 of 0.50, while the optimal (10, 5) setting achieved MRR@5 of 0.750 with perfect Recall@5 of 1.00—a 59\% absolute MRR improvement. All tested reranking settings achieved perfect recall (1.00), indicating improved ranking quality and coverage.

Large initial retrieval depths ($k_{initial} \geq 50$) showed diminishing returns. The optimal (10, 5) setting balances ranking quality with reasonable latency (2.02 seconds), while (10, 10) provides lowest latency (1.24 seconds) for sub-2-second response requirements. Our evaluation focused on Cohere \texttt{rerank-english-v3.0}, comparing alternative reranking architectures including LLM-based rerankers and domain-specific financial rerankers would provide additional insights into the accuracy-latency tradeoffs for financial document retrieval.

\subsection{Small-to-Big Retrieval}

Small-to-big retrieval achieved a 65\% win rate with only 0.2 seconds latency increase, demonstrating that augmenting target chunks with neighboring context improves answer generation with negligible overhead. The asynchronous implementation (0.17 seconds) outperformed synchronous retrieval (0.34 seconds) through parallel database queries. Cost analysis revealed zero per-query cost increase (\$0.000078). 

The effectiveness of small-to-big retrieval depends on whether surrounding sentences provide relevant context. For needle-in-haystack queries where the answer is a single isolated fact, expanding context offers limited benefit. However, for multi-hop reasoning queries requiring comprehensive context spanning chunk boundaries, small-to-big retrieval substantially improves answer quality by providing the LLM with richer surrounding information. These results suggest small-to-big retrieval offers substantial benefits for production systems handling queries that benefit from expanded context windows. Our implementation expanded each target chunk with its immediate neighbors (one preceding and one following chunk). Future work can explore how different expansion window sizes effect answer quality and context distraction during generation.

\section{\uppercase{Conclusion}}\label{sec:conclusion}

As financial question answering systems scale to lengthy SEC filings spanning hundreds of pages, balancing retrieval precision, answer quality, and computational efficiency remains essential. We presented a systematic evaluation comparing vector-based agentic RAG with hybrid search and metadata filtering against hierarchical node-based reasoning systems, additionally evaluating cross-encoder reranking and small-to-big chunk retrieval as independent enhancement techniques. Across 1,200 U.S. SEC filings (10-K, 10-Q, and 8-K) on a 150-question benchmark, vector-based agentic RAG achieved a 68\% win rate over hierarchical node-based systems with comparable latency (5.2 compared to 5.98 seconds), with the hierarchical system's bottleneck emerging at the table-of-contents selection stage where the LLM struggled to identify relevant sections. Cross-encoder reranking delivered substantial improvements over baseline vector retrieval, increasing Mean Reciprocal Rank from 0.160 to 0.750 (59\% absolute improvement) with perfect Recall@5 (1.00), demonstrating clear gains in information retrieval metrics. Small-to-big retrieval achieved a 65\% win rate with only 0.2 seconds additional latency, confirming that expanded context improves answer quality when surrounding sentences contain relevant information. 

\bibliographystyle{apalike}
{\small
\bibliography{references}}



\end{document}